\documentclass[10pt,twocolumn,letterpaper]{article}

\usepackage{cvpr}              
\usepackage{multirow}
\usepackage{hyperref}       
\usepackage{graphicx}
\usepackage{algorithm}
\usepackage{algorithmic}
\usepackage{subcaption}
\usepackage[dvipsnames]{xcolor}

\definecolor{cvprblue}{rgb}{0.21,0.49,0.74}
\def\paperID{*****} 
\def\confName{CVPR}
\def\confYear{2024}

\title{MedSAGa: Few-shot Memory Efficient
Medical Image Segmentation using Gradient
Low-Rank Projection in SAM}

\author{Navyansh Mahla $^*$\\ 
Indian Institute of Technology, Bombay\\
Mumbai, Maharashtra, India\\
{\tt\small 210040106@iitb.ac.in}
\and
Annie D'souza $^*$\\
Indian Institute of Technology, Bombay\\
Mumbai, Maharashtra, India\\
{\tt\small 
20d070028@iitb.ac.in}
\and
Shubh Gupta\\
Sardar Patel Institute of Technology, Mumbai\\
Mumbai, Maharashtra, India\\
{\tt\small shubh.gupta@spit.ac.in}
\and
Bhavik Kanekar\\
Indian Institute of Technology, Bombay\\
Mumbai, Maharashtra, India\\
{\tt\small 
bhavikkanekar@iitb.ac.in}
\and
Kshitij Sharad Jadhav\\
Indian Institute of Technology, Bombay\\
Mumbai, Maharashtra, India\\
{\tt\small kshitij.jadhav@iitb.ac.in}
}

\begin{document}
\maketitle
\def\thefootnote{*}\footnotetext{These authors contributed equally to this work}\def\thefootnote{\arabic{footnote}}
\begin{abstract}
The application of large-scale models in medical image segmentation demands substantial quantities 
of meticulously annotated data curated by experts along with high computational resources, both of which are challenges in resource-poor settings. In this study, we present the Medical Segment Anything Model with Galore \textbf{(MedSAGa)} where we adopt
the Segment Anything Model (SAM) to achieve memory-efficient, few-shot medical image segmentation by applying Gradient Low-Rank Projection (\textbf{GaLore}) to the parameters of the image encoder of SAM. Meanwhile, the weights of the prompt encoder and mask decoder undergo full parameter fine-tuning using standard optimizers. We further assess MedSAGa's few-shot learning capabilities, reporting on its memory efficiency and segmentation performance across multiple standard medical image segmentation datasets. We compare it with several baseline models, including LoRA fine-tuned SAM (SAMed) and DAE-Former. Experiments across multiple datasets and these baseline models with different number of images for fine tuning demonstrated that the GPU  memory consumption of MedSAGa is significantly less than that of the baseline models, \textbf{achieving an average memory efficiency of 66\%} more than current state-of-the-art (SOTA) models for medical image segmentation. The combination of substantially lower memory requirements and comparable to SOTA results in few-shot learning for medical image segmentation positions MedSAGa as an optimal solution for deployment in resource-constrained settings.
\end{abstract}    
\section{Introduction}
\label{sec:intro}

Image segmentation plays an important role in various aspects of healthcare, enabling precise analysis and diagnosis from medical imaging data such as MRI, CT scans, and ultrasound \cite{wang2022medical}. By accurately delineating anatomical structures or pathological regions, medical image segmentation could assist clinicians in tracking the health of the patients by identifying abnormalities and planning treatments \cite{zhang2024segment,ma2024segment}. This is performed by clinical experts who manually outline the borders for segmentation which can subsequently be used in deep learning algorithms. However, labeling medical images requires consensus of multiple clinical experts making it expensive and difficult, especially in resource-constrained settings. Few-shot learning and zero-shot learning prove to be very useful in such scenarios \cite{LU202225}.\par 
Over the past decade, a multitude of deep learning models, including U-Net \cite{unet10.1007}, and transformer-based models such as TransUNet \cite{chen2021transunet} and DAE-Former \cite{azad2023dae}, have been developed for image segmentation tasks. The latest large-scale models (LM) such as GPT-4 \cite{achiam2023gpt}, SAM \cite{Kirillov_2023_ICCV}, DALL-E \cite{ramesh2021zero} and SegGPT \cite{Wang_2023_ICCV} provided a platform to solve different image segmentation tasks. These models are trained on huge datasets and the performance of these models is highly competitive even in zero-shot learning.  However, resources such as memory and the compute required for training and fine-tuning these models for downstream tasks are significantly large making it difficult to deploy them in a resource-constrained setting. Although the SAM and SegGPT models show SOTA performances, these models are not trained on medical images and thus, cannot be utilized off-the-shelf for tasks like medical image segmentation. Hence, an efficient fine-tuning strategy is required to utilize the above-mentioned large-scale models for downstream tasks like medical image segmentation is required. There are several Parameter Efficient Fine-Tuning (PEFT) strategies which are categorized into additive, selective, reparameterized, and hybrid fine-tuning based on their operations \cite{zhang2023customized}. In this work, we utilize \textbf{SAM} for the medical image segmentation task and adopt the Gradient Low-rank projection (\textbf{GaLore}) strategy to fine-tune it on medical image segmentation data \cite{zhao2024galore}.  We summarize our contributions in the following points:
\begin{enumerate}
    \item We demonstrate \textbf{Med}ical \textbf{S}egment \textbf{A}nything Model with \textbf{Ga}Lore (\textbf{MedSAGa}), a framework integrating the Gradient Low-rank Projection (GaLore) optimization with SAM.
    \item We perform rigorous experimentation on four diverse medical image segmentation datasets and compare the performance of MedSAGa with standard benchmarks in few-shot settings.
    \item Our results showcase the significance of our framework by demonstrating notable reductions in memory over existing models while delivering comparable performance.
\end{enumerate}

\begin{figure}[ht]
    \centering
    \begin{minipage}{0.48\textwidth}
        \centering
        \includegraphics[width=\linewidth]{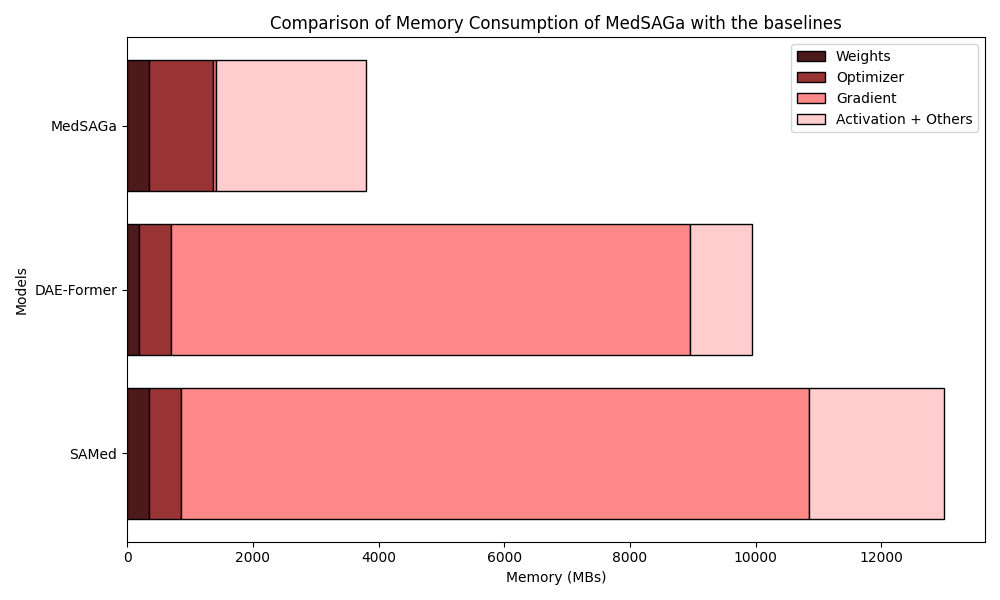} 
        \caption{Memory consumption of MedSAGa and the other standard baselines while fine-tuning them for medical image segmentation task.}
        \label{fig:memory_graph}
    \end{minipage}
    \hfill 
    \begin{minipage}{0.48\textwidth}
        \centering          
        \includegraphics[width=2.3in]{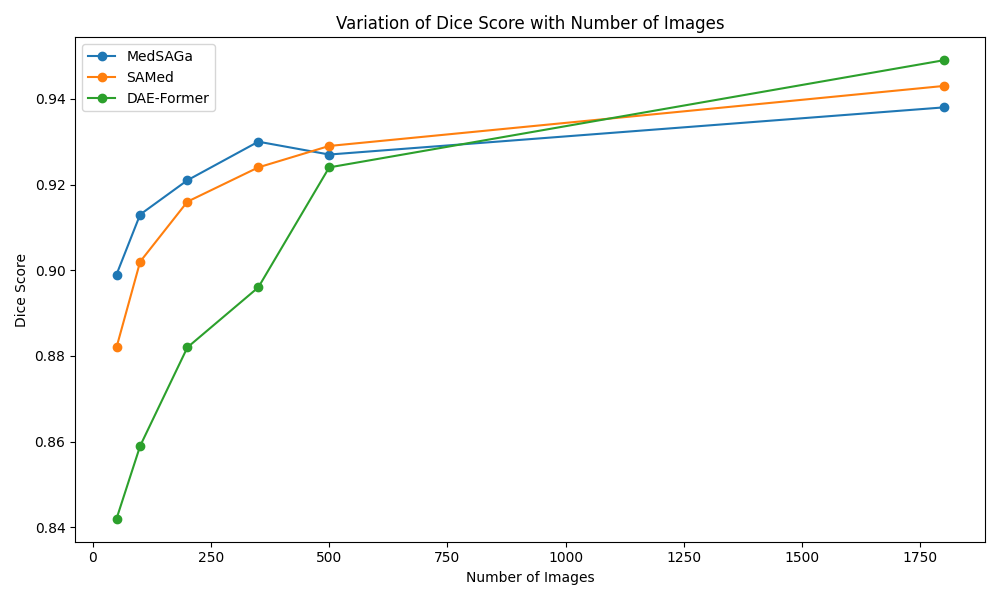}
        \caption{Dice Score vs Number of images while fine-tuning on ChestX-ray8 dataset. Here, for most models, the graph plateaus out at approximately 500 images.}
        \label{fig:few_shot_graph}
    \end{minipage}
\end{figure}

\begin{figure}
    \centering
        \includegraphics[width =\linewidth, height=3.8cm]{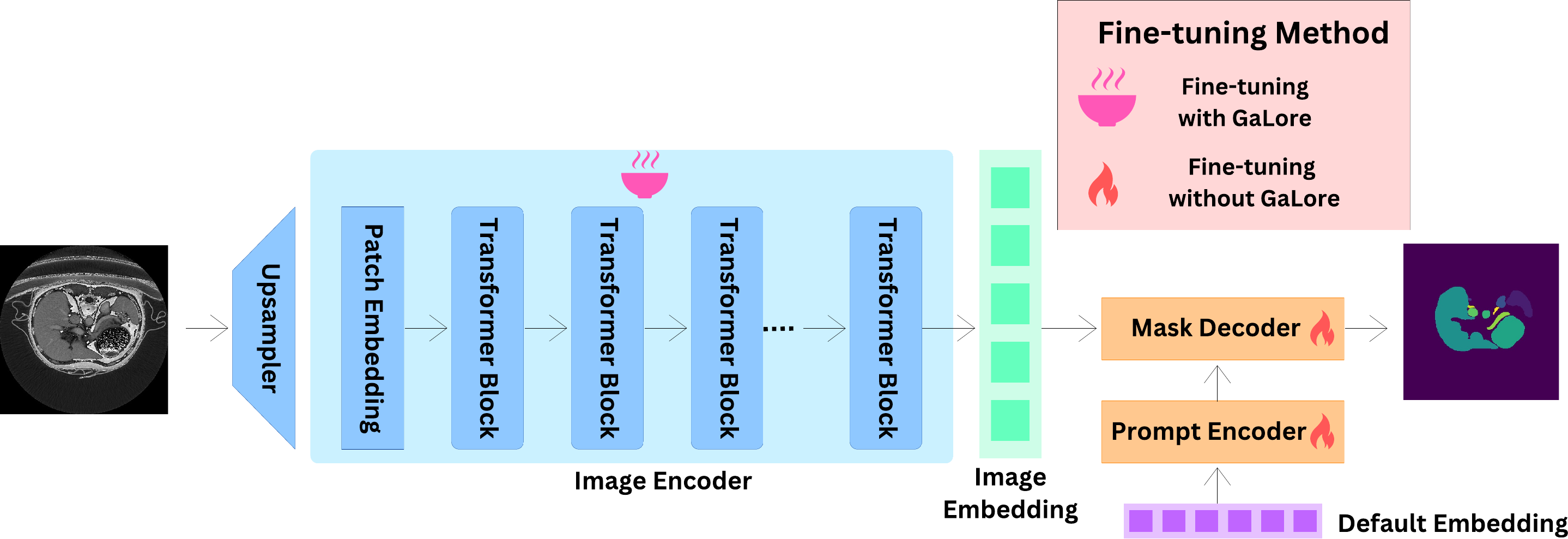}
        \caption{The architecture of MedSAGa. GaLore optimization is applied to fine-tune only the image encoder. Due to their lightweight characteristics, the Mask Decoder and the default embeddings of the Prompt Encoder are fine-tuned directly on the medical images without applying GaLore.}
        \label{fig:Our apporach}
\end{figure}

\section{Related Works}
\label{sec:formatting}

 A major milestone for medical image segmentation was achieved by developing U-Net, a model based on convolutional neural networks. Its novel architecture provides precise medical image segmentation even when trained on a limited number of images \cite{unet10.1007}. Further, several different variations of U-Net have been developed like DenseUNet \cite{cao2020denseunet} and ResUNet \cite{rahman2022ResUnet}, which improved the segmentation performance by making significant changes in the structure of the U-Net architecture. The TransUNet model proposed in study \cite{chen2021transunet} utilizes the transformer architecture for encoding in conjunction with the U-Net architecture \cite{castro2022u}. It benefits from the global contextual information reception capabilities of transformers, that U-Net’s convolutional layers lack due to their local receptive fields. However, transformers require substantial computational resources, particularly in terms of GPU memory, because they process the entire image as a sequence of patches \cite{dosovitskiy2021an}. This sequential processing leads to high memory demands, especially when handling larger images typical in medical imaging applications. The hybrid nature of TransUNet, which combines convolutional operations with transformer mechanisms, increases the complexity of the model. This complexity can make the training process more computationally intensive \cite{chen2021transunet}. Further, the transformer layers in TransUNet require multiple self-attention calculations which are computationally expensive, especially for higher resolution inputs \cite{chen2021transunet}. Due to the complexity and larger number of parameters, training TransUNet can be time and resource-consuming. This is exacerbated when fine-tuning on specific tasks as transformers generally take longer to train than their purely convolutional counterparts. The use of advanced data augmentation and training strategies necessary to achieve optimal performance further extends the training duration and computational expense of TransUNet \cite{chen2021transunet}. An architecture based on transformers, called DAE-Former \cite{azad2023dae}, was further proposed, which utilizes an efficient dual attention mechanism and has been demonstrated to be better than TransUNet in terms of both computational efficiency and accuracy. However, the self-attention mechanism of DAE-Former, although redesigned for efficiency, still retains the intrinsic quadratic computational complexity with respect to the number of tokens. This complexity can lead to significant computational overhead when processing large datasets typical in medical imaging \cite{azad2023dae}.  Thus, although these models prove to be quite efficient in image segmentation when trained for a particular task, for a diverse and resource-constrained domain such as healthcare, a generalized large-scale model capable of few-shot learning should be the most practical framework.

In the domain of large-scale free lunch models for image segmentation, SAM (Segment Anything Model) represents a significant milestone, trained on an extensive corpus of over 1B masks from 11M images \cite{Kirillov_2023_ICCV}. It is noted for its superior performance, especially in zero-shot settings where it performs similarly to or even better than many supervised models. However, SAM doesn't perform well on medical image segmentation as it is trained on natural images and hence, cannot capture the semantics required in medical image segmentation \cite{leng2024self}. After fine-tuning the SAM model for medical image segmentation task, SAM shows promising few shot learning results \cite{mazurowski2023segment}. 

Parameter Efficient Fine-tuning (PEFT) techniques provide a solution for adapting such large-scale models for downstream tasks by reducing the number of trainable parameters while maintaining comparable performance \cite{xu2023parameter}. One such approach, known as LoRA (Low-Rank Adaptation) does this by freezing the pre-trained model weights and integrating low-rank matrices into the transformer layers \cite{hu2021lora}. This significantly reduces the number of trainable parameters required for fine-tuning the large-scale models for various downstream tasks. A medical image segmentation model, SAMed (Customized Segment Anything Model for
Medical Image Segmentation) utilizes these LoRA layers for fine-tuning the image encoder of SAM \cite{zhang2023customized}. It further updates the prompt encoder and mask decoder parameters on medical image segmentation datasets which enhances the segmentation capability of SAM on medical images. This approach provides significant improvement in performance while utilizing less memory for fine-tuning on medical images \cite{zhang2023customized}.

A novel approach to reducing memory consumption was introduced in \textbf{GaLore} (Gradient Low-Rank Projection) \cite{zhao2024galore} which presents a memory-efficient technique for fine-tuning transformer-based models by projecting gradients into a low-rank space during training. This method effectively reduces the memory footprint of optimizer states while preserving the capability to learn full-rank weights. As a result, GaLore enables more efficient training on consumer-grade GPUs without the necessity for model parallelism or checkpointing. In our framework, MedSAGa, we leverage GaLore for fine-tuning the SAM architecture, showcasing its effectiveness in medical image segmentation, particularly in resource-constrained environments.

\section{Methodology}

\subsection{Overview}
Through MedSAGa, our primary aim is to harness the few-shot capabilities of a large-scale image segmentation model such as SAM and apply it to a resource-constrained setting like healthcare \cite{bernhardt2022active}. As seen in fig.~\ref{fig:memory_graph}, MedSAGa is more efficient than most SOTA models in both the training as well as inferencing stage (which uses \textit{weights+activations+others} memory). By resource-constrained settings, we refer to the requirement of a substantially less number of annotated data to train or fine-tune the segmentation models and an inability to procure huge memory and computational resources required to train large-scale models. SAMed attempted to address this by integrating LoRA layers into its image encoder \cite{zhang2023customized}, however, it still requires huge memory and computational resources while facing the challenge of not utilizing the entire parameter matrix for training. Hence, in our work \textbf{MedSAGa}, we demonstrate a more refined approach to adapt SAM in resource-constrained settings by applying \textbf{GaLore} to all the parameters of the image encoder and then fine-tuning it, which significantly reduces the memory and computational cost while still maintaining full parameter structure. In addition to this, we fine-tune the prompt encoder and mask decoder without GaLore to perform improved semantic segmentation on medical images, as is demonstrated in SAMed \cite{zhang2023customized}. We note here that since the prompt encoder and mask decoder in SAM are lightweight \cite{zhang2023customized}, applying Galore for fine-tuning them would not lead to a substantial improvement in memory consumption. Fig.~\ref{fig:memory_graph} shows the memory utilization by MedSAGa and the other SOTA models while fine-tuning on medical images. Furthermore, as seen in fig.~\ref{fig:few_shot_graph}, it can be observed that after a specific number of images, the performance graph plateaus out supporting the idea to use few-shot learning in resource constrained settings giving very close performance to using the entire dataset while using a smaller proportion of images.

SAM generates multiple segmentation masks to avoid ambiguity. However, we align MedSAGa with the working of SAMed and generate multiple segmentation masks, each representing a different tissue or segment of the anatomy in addition to the background mask.
These masks are then further post-processed to give the final segmentation result. For the training phase, we adopt warmup to stabilize the training process and use the AdamW optimizer for improved performance as was suggested in the SAMed architecture \cite{zhang2023customized}.

\subsection{The Architecture}

To reduce the resources required for utilizing SAM, we apply GaLore optimization to all the parameters of the image encoder. This approach harnesses the gradually changing low-rank structure of the gradient explained in the GaLore paper, which improves memory efficiency while still giving comparable results \cite{zhao2024galore}. Instead of reducing the weight parameter size as is done in LoRA, GaLore projects the Gradient matrix at time \( t \), \( G_t \in \mathbb{R}^{m \times n} \), into a low-rank matrix \( \tilde{G}_t \), which can be represented by eq. \ref{eq-1}.
\begin{equation}
\quad \tilde{G}_t = P_t \rho_t(P_t^\top G_t Q_t) Q_t^\top. \\
\label{eq-1}
\end{equation}  
where \(P_t\) and \(Q_t\) are projection matrices with dimensions \(\mathbb{R}^{m \times r}\) and \(\mathbb{R}^{n \times r}\) respectively, and \(\rho_t\) is an element-wise stateful gradient regularizer. If \( W_0 \) is the initial weight matrix, \( W_T \) represents the weight matrix at time \( T \), and \( \eta \) is the learning rate, then the gradient update rule in GaLore is as follows in eq. \ref{eq-2}.
\begin{equation}
W_T = W_0 + \eta \sum_{t=0}^{T-1} \tilde{G}_t.
\label{eq-2}
\end{equation}

In our approach, we apply this Gradient low-rank projection to all the parameters of the image encoder which includes all the projection layers (\textit{q, k, v and o}) as opposed to SAMed which applies LoRA only to the \textit{q} and \textit{v} projection layers in its best-performing model. We apply GaLore to all the parameters of the Image encoder as it adds only a negligible memory overhead. For the MedSAGa approach to function as an auto-segmentation model, we do not provide any prompts to the prompt encoder. Instead, we utilize the default embedding of the prompt encoder of SAM, which it uses when no prompt is given, and only fine-tune it during the training phase.

The mask decoder of SAM consists of a lightweight transformer layer and a segmentation head \cite{zhang2023customized}. In our approach, we fine-tune the entire mask decoder directly without applying any optimization as it is already lightweight. Furthermore, as was developed in the SAMed architecture, we change the segmentation head of the mask decoder to adapt it to give precise semantic segmentation for each class of tissue or anatomy present in the image. Let us consider there are \( k \) classes in total including 1 background class in the medical image, the mask decoder of MedSAGa predicts \( k \) segmentation masks \( M \in \mathbb{R}^{h \times w \times k} \), each corresponding to a single class in the image. We then further utilize a combination of \textit{argmax} and \textit{softmax} functions to generate a segmentation map \( \hat{Y} \) as shown in eq. \ref{eq-3} where $d = -1$ represents the channel dimensions..
\begin{equation}
\hat{Y} = \text{argmax}(\text{Softmax}(M, d = -1), d = -1)
\label{eq-3}
\end{equation}

These adjustments make MedSAGa is an easy to implement solution in the SAM architecture with minimal engineering required, thereby making our solution very practical and adaptable for varied settings.

\subsection{Training Strategies}
For performing training, we utilize a combination of Cross Entropy loss and Dice loss, similar to that utilized in the SAMed approach as represented by the eq.~\ref{eq-4}.
\begin{equation}
    \mathcal{L}=\lambda\mathcal{L}_{CE}(M, D(S))+(1-\lambda)\mathcal{L}_{\text{Dice}}(M, D(S))
    \label{eq-4},
\end{equation}
where \(\mathcal{L}\) represents the net loss value, \(\mathcal{L}_{CE}\) represents the Cross Entropy loss, \(\mathcal{L}_{\text{Dice}}\) represents the Dice loss  and \( \lambda \) represents the loss weight. $D$ represents downsampling to align the resolution of the ground truth mask ($S$) with the MedSAGa output, compensating for lower spatial resolution of MedSAGa.

Warmup is applied in MedSAGa to stabilize the training process. By allowing the learning rate to increase gradually, we enable the model to slowly adapt the weights to the specific characteristics of the medical data, thereby avoiding poor convergence and instability early in the training phase and reducing the chances of overfitting \cite{xiong2020layer}.

Instead of Adam optimizer, in MedSAGa we utilize the AdamW optimizer. AdamW decouples weight decay from the gradient updates and applies it directly to the weights. This method proves to be more effective in regularization, maintaining a better separation between the weight decay and the adaptive learning rate aspects of Adam \cite{loshchilov2018decoupled}. In MedSAGa, the AdamW approach ensures that the regularization is not overly influenced by the learning rate adaptations specific to different weights.


\section{Experiments and Results}

We demonstrate the performance of MedSAGa through rigorous experimentation on 4 different medical datasets by comparing it to several baseline models.

\subsection{Datasets and Evaluation Metrics}
\textbf{Datasets.} We utilize four different datasets covering different parts of the human anatomy for our experimentation. All the baselines and results that we present in the further sections have been tested on each of these 4 datasets to evaluate the robustness of the MedSAGa architecture. For each dataset, the number of few-shot images used for experimentation are chosen depending upon the size of the dataset and the number of classes. 

The \textit{AMOS dataset} is a large-scale clinical dataset of 500 MRI and 100 CT scans which consists of annotations of 15 abdominal organs \cite{ji2022amos}. Each slice was padded to obtain final slices of dimension 512 \(\times\) 512. For the training set, we only considered the slices that had masks of at least 5 organs to overcome class imbalance. In total, 10,300 slices were satisfying the above criteria out of which we used 500, 1000, 2000, 5000, and 7000 slices for the few shot experimentation. The \textit{ChestX-ray8 dataset} consists of 108,948 frontal-view X-ray images of 32,717 unique patients \cite{wang2017chestx}. The images consist of 8 labels, mined from the text corpus of the corresponding radiological reports using NLP. We predict the segmentations for three classes whose masks were available: left lung, right lung and heart. Out of the 108,948 images in the dataset, we used only 50, 100, 200, 350, 500 and 1800 images as performance saturation was reached at quite an early stage (Refer Fig.~\ref{fig:few_shot_graph}). The \textit{Ischemic Stroke Lesion Segmentation (ISLES) dataset} is a multi-center MRI dataset of acute to subacute stroke lesions \cite{hernandez2022isles}. It consists of 400 multi-vendor MRI images, each consisting of a number of slices having a dimension of 112 \(\times\) 112. The DWI modality was used for experimentation and as a pre-processing step, thresholding was applied to only use slices containing lesions while training. For performing few-shot experiments, 200, 500, 700, 1000, 2000 and 3,200 (maximum number of slices obtained after thresholding) slices were used in the ISLES dataset. The \textit{Spleen dataset} was retrieved from the Medical Segmentation Decathlon challenge \cite{antonelli2022medical}. It consists of 61 3D volume portal-venous phase CT scans from patients undergoing treatment for liver metastases. Out of these, we utilize only those images for training that have a mask available. 
Data leakage between the train and test sets was avoided by splitting the data according to the patients instead of according to individual slices for all datasets. \\
\textbf{Evaluation Metrics} We use dice score and HD95 metrics to compare the model performances. Also, we show the memory utilized by the model for training and fine tunning. 

\begin{table}[]
\caption{Results of segmentation performances across datasets}
\label{tab:my-table1}
\begin{tabular}{ccccccc}
\hline
\multicolumn{1}{c|}{\multirow{2}{*}{\textbf{\begin{tabular}[c]{@{}c@{}}No. of \\ Images\end{tabular}}}} & \multicolumn{2}{c|}{\textbf{MedSAGa}}               & \multicolumn{2}{c|}{\textbf{SAMed}}                 & \multicolumn{2}{c}{\textbf{DAE-Former}} \\ \cline{2-7} 
\multicolumn{1}{c|}{}                                                                                   & \textit{Dice}  & \multicolumn{1}{c|}{\textit{hd95}} & \textit{Dice}  & \multicolumn{1}{c|}{\textit{hd95}} & \textit{Dice}       & \textit{hd95}     \\ \hline
\multicolumn{7}{c}{\textbf{Chest-Xray Dataset}}                                                                                                                                                                                                               \\ \hline
\multicolumn{1}{c|}{\textit{50}}                                                                        & \textbf{0.899} & \multicolumn{1}{c|}{44.704}        & 0.882          & \multicolumn{1}{c|}{48.772}        & 0.842               & 183.220           \\
\multicolumn{1}{c|}{\textit{100}}                                                                       & \textbf{0.913} & \multicolumn{1}{c|}{39.430}        & 0.902          & \multicolumn{1}{c|}{39.361}        & 0.859               & 130.535           \\
\multicolumn{1}{c|}{\textit{200}}                                                                       & \textbf{0.921} & \multicolumn{1}{c|}{33.745}        & 0.916          & \multicolumn{1}{c|}{34.372}        & 0.882               & 98.342            \\
\multicolumn{1}{c|}{\textit{350}}                                                                       & \textbf{0.930} & \multicolumn{1}{c|}{29.235}        & 0.924          & \multicolumn{1}{c|}{30.421}        & 0.896               & 72.783            \\
\multicolumn{1}{c|}{\textit{500}}                                                                       & \textbf{0.934} & \multicolumn{1}{c|}{30.502}        & 0.929          & \multicolumn{1}{c|}{27.764}        & 0.924               & 43.698            \\
\multicolumn{1}{c|}{\textit{1800}}                                                                      & 0.938          & \multicolumn{1}{c|}{24.456}        & 0.943          & \multicolumn{1}{c|}{21.652}        & \textbf{0.949}      & 19.986            \\ \hline
\multicolumn{7}{c}{\textbf{ISLES Dataset}}                                                                                                                                                                                                                    \\ \hline
\multicolumn{1}{c|}{\textit{200}}                                                                       & \textbf{0.343} & \multicolumn{1}{c|}{11.988}        & 0.218          & \multicolumn{1}{c|}{26.736}        & 0.267               & 14.296            \\
\multicolumn{1}{c|}{\textit{500}}                                                                       & \textbf{0.537} & \multicolumn{1}{c|}{12.840}        & 0.227          & \multicolumn{1}{c|}{22.560}        & 0.435               & 19.890            \\
\multicolumn{1}{c|}{\textit{700}}                                                                       & 0.534          & \multicolumn{1}{c|}{11.011}        & 0.280          & \multicolumn{1}{c|}{20.140}        & \textbf{0.553}      & 13.919            \\
\multicolumn{1}{c|}{\textit{1000}}                                                                      & \textbf{0.574} & \multicolumn{1}{c|}{8.900}         & 0.368          & \multicolumn{1}{c|}{20.690}        & 0.546               & 12.774            \\
\multicolumn{1}{c|}{\textit{2000}}                                                                      & 0.685          & \multicolumn{1}{c|}{6.953}         & \textbf{0.686} & \multicolumn{1}{c|}{6.953}         & 0.683               & 9.962             \\
\multicolumn{1}{c|}{\textit{3200}}                                                                      & \textbf{0.783} & \multicolumn{1}{c|}{3.748}         & 0.742          & \multicolumn{1}{c|}{14.977}        & 0.733               & 11.019            \\ \hline
\multicolumn{7}{c}{\textbf{Spleen Dataset}}                                                                                                                                                                                                                   \\ \hline
\multicolumn{1}{c|}{\textit{100}}                                                                       & \textbf{0.842} & \multicolumn{1}{c|}{14.645}        & 0.807          & \multicolumn{1}{c|}{15.836}        & 0.793               & 20.029            \\
\multicolumn{1}{c|}{\textit{200}}                                                                       & 0.838          & \multicolumn{1}{c|}{13.598}        & 0.869          & \multicolumn{1}{c|}{10.861}        & \textbf{0.877}      & 39.166            \\
\multicolumn{1}{c|}{\textit{500}}                                                                       & 0.882          & \multicolumn{1}{c|}{12.006}        & 0.877          & \multicolumn{1}{c|}{10.473}        & \textbf{0.905}      & 26.740            \\
\multicolumn{1}{c|}{\textit{700}}                                                                       & 0.876          & \multicolumn{1}{c|}{12.176}        & 0.878          & \multicolumn{1}{c|}{4.643}         & \textbf{0.901}      & 12.290            \\
\multicolumn{1}{c|}{\textit{800}}                                                                       & 0.882          & \multicolumn{1}{c|}{15.851}        & 0.907          & \multicolumn{1}{c|}{49.979}        & \textbf{0.916}      & 26.016            \\ \hline
\multicolumn{7}{c}{\textbf{AMOS Dataset}}                                                                                                                                                                                                                     \\ \hline
\multicolumn{1}{c|}{\textit{500}}                                                                       & 0.199          & \multicolumn{1}{c|}{66.697}        & 0.166          & \multicolumn{1}{c|}{71.604}        & \textbf{0.268}      & 101.784           \\
\multicolumn{1}{c|}{\textit{1000}}                                                                      & 0.208          & \multicolumn{1}{c|}{64.047}        & 0.201          & \multicolumn{1}{c|}{68.203}        & \textbf{0.301}      & 93.830            \\
\multicolumn{1}{c|}{\textit{2000}}                                                                      & 0.222          & \multicolumn{1}{c|}{68.311}        & 0.201          & \multicolumn{1}{c|}{67.397}        & \textbf{0.299}      & 92.947            \\
\multicolumn{1}{c|}{\textit{5000}}                                                                      & \textbf{0.314} & \multicolumn{1}{c|}{79.729}        & 0.250          & \multicolumn{1}{c|}{62.494}        & \textbf{0.314}      & 85.232            \\
\multicolumn{1}{c|}{\textit{7000}}                                                                      & \textbf{0.389} & \multicolumn{1}{c|}{52.768}        & 0.377          & \multicolumn{1}{c|}{56.124}        & 0.349               & 79.323            \\ \hline
\end{tabular}
\end{table}

\subsection{Implementation details and evaluation metrics}
All the experiments were run on a single NVIDIA RTX A6000 GPU with 48GB GPU RAM. A batch size of 12 was used for all the experiments. The warmup and loss weight strategies were the same as used in SAMed, i.e., the loss weights for cross entropy and dice loss were set to 0.2 and 0.8, respectively. For the warmup, the initial learning rate was set to 0.005, while the warmup period was set to 250. The $\beta_1$, $\beta_2$, and the weight decay of the AdamW optimizer were set to 0.9, 0.999, and 0.1, respectively. For the GaLore optimizer, AdamW was used as the base optimizer with a learning rate of $1 \times 10^{-3}$, and $\beta_1$ and $\beta_2$ were set to 0.9 and 0.999, respectively. The plane change rate $(T)$ used was 200, as is recommended in the GaLore Paper. The base pre-trained model architecture for training our SAM-based model was the \texttt{vit\_b} model, which occupies 5.25\% (18.81M) of the original model size (358M).
\begin{table}[htbp]  
\centering
\caption{Comparison of segmentation performance of MedSAGa on the Spleen dataset, when fine-tuned with and without using warmup on Chest X-ray dataset.}
\label{tab:my-table}
\resizebox{\linewidth}{!}{
\begin{tabular}{c|cc|cc}

\hline
\multirow{2}{*}{\textbf{Dataset}} & \multicolumn{2}{c|}{\textbf{With Warmup}} & \multicolumn{2}{c}{\textbf{Without warmup}} \\
          & \textit{Mean Dice}  & \textit{Mean HD95}  & \textit{Mean Dice}   & \textit{Mean HD95}   \\ \hline
\textit{Spleen (500img)}          & 0.882               & 12.006              & 0.858                & 18.731               \\
\textit{ISLES(200img)}            & 0.343               & 11.988              & 0.083                & 12.112               \\
\textit{CXR(100img)}              & 0.913               & 39.430              & 0.736                & 109.502              \\
\textit{AMOS(500img)}             & 0.199               & 66.697              & 0.196                & 68.014               \\ \hline
\end{tabular}

}

\end{table}


\begin{figure}[htbp]  

\centering
\includegraphics[width=\linewidth]{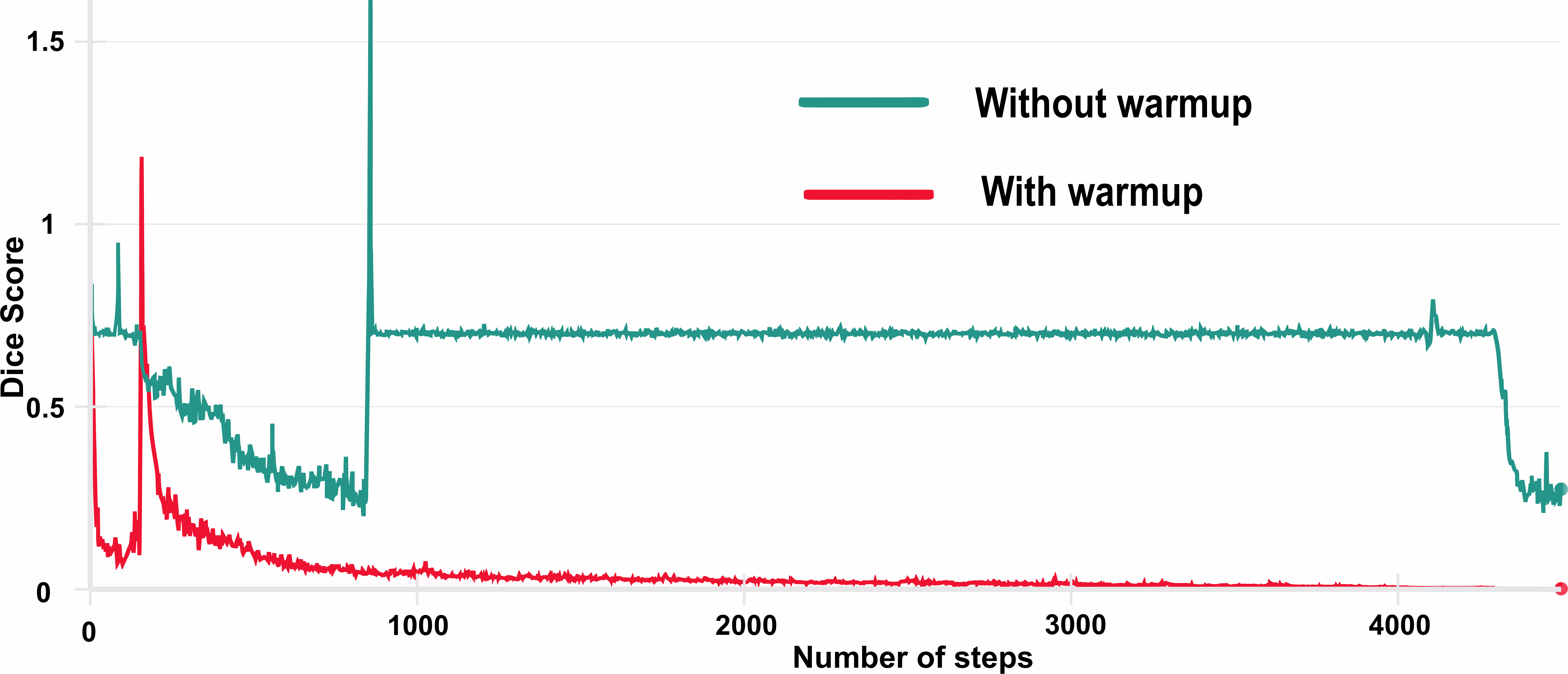} 
\caption{Comparison of loss curves between fine-tuning with and without applying warmup on ChestX-ray8 dataset.}
\label{fig:my-figure}
\end{figure}


\subsection{Comparison with SOTA}
The results of the memory utilization by MedSAGa and other baselines are depicted in Fig.~\ref{fig:memory_graph}. From this it is evident that the MedSAGa model utilizes the lowest memory as compared to the other SOTA image segmentation models, giving an average memory efficiency of 66\% more
as compared to current state-of-the-art (SOTA) models for medical image segmentation. Here, we would like to mention that since the memory utilization reported in the Fig.~\ref{fig:memory_graph} are the model memories, they are independent of the training data and depend only on the model parameters. Hence, MedSAGa is highly memory efficient as compared to the other standard baselines in both training and inferencing phases.

The results of the segmentation performance are compiled in the Table~\ref{tab:my-table1}. The table shows the performance metrics, the mean Dice Scores and the mean HD95 scores, for various few-shot settings across the four medical image datasets. These results depict the ability of \textbf{MedSAGa} as the best approach in terms of memory efficiency while still giving comparable segmentation performances in most of the settings. In the domain of SAM-based models for medical image segmentation, we beat the state-of-the-art model, SAMed not only in terms of memory efficiency but also in segmentation performance in almost every few shot settings across all datasets.

Comparing with DAE-Former, MedSAGa performs better segmentation in very low resource settings for most datasets like Chest X-ray, ISLES and Spleen datasets with a slight drop in the dice score when higher number of training images are used. Furthermore, MedSAGa utilizes 61\% less memory compared to DAE-Former. 


\begin{table}[]
\centering
\caption{Ablation Studies of variations in fine-tuning SAM architecture with GaLore}
\label{tab:ablation-table}
\resizebox{1.0\linewidth}{!}{
\begin{tabular}{c|c|cc|cc|cc}
\toprule
\multirow{2}{*}{\textbf{Dataset}}                                          & \multirow{2}{*}{\textbf{\begin{tabular}[c]{@{}c@{}}No. of \\ Images\end{tabular}}} & \multicolumn{2}{c|}{\textbf{Medsaga}}                   & \multicolumn{2}{c|}{\textbf{Medsaga\_v1}}                & \multicolumn{2}{|c}{\textbf{Medsaga\_v2}}        \\ 
\cmidrule(lr){3-4} \cmidrule(lr){5-6} \cmidrule(lr){7-8}
&      
& \textit{Dice}             & \textit{hd95}              & \textit{Dice}             & \textit{hd95}               & \textit{Dice}             & \textit{hd95}               \\

\midrule

\multirow{2}{*}{\textit{\begin{tabular}[c]{@{}c@{}}Chest-\\ Xray\end{tabular}}} & 50      & 0.899     & 44.704         & 0.703    & 138.368    & 0.119    & 492.937      \\
& 1800       & 0.943      & 24.456      & 0.876  & 57.477     & 0.119      & 492.937      \\
\midrule

\multirow{2}{*}{\textit{ISLES}}  & 200    & 0.343   & 11.988        & 0.415     & 16.011    & 0.007    & 59.111        \\
& 3200    & 0.783     & 3.748   & 0.783    & 3.748   & 0.007          & 59.111   \\

\midrule

\multirow{2}{*}{\textit{Spleen}} & 100  & 0.842 & 14.645   & 0.439    & 30.209  & 0.006   & 295.004      \\
& 800     & 0.882   & 15.851  & 0.471  & 29.551 & 0.006   & 295.004   \\  
\midrule
\multirow{2}{*}{\textit{AMOS}}  & 500   & \multicolumn{1}{r}{0.199} & \multicolumn{1}{r}{66.696} & \multicolumn{1}{|r}{0.198} & \multicolumn{1}{r}{90.752}  & \multicolumn{1}{|r}{0.016} & \multicolumn{1}{r}{231.947} \\
& 7000   & \multicolumn{1}{r}{0.359} & \multicolumn{1}{r}{52.768} & \multicolumn{1}{|r}{0.325} & \multicolumn{1}{r}{90.058} & \multicolumn{1}{|r}{0.016} & \multicolumn{1}{r}{231.947} \\    
\bottomrule
\end{tabular}
}
\end{table}

\section{Ablation Studies}
We present our ablation studies in two categories. In the first part, we demonstrate two variations of integrating Gradient low-rank projection optimization in the MedSAGa architecture. In the first variation (referred to as \textit{MedSAGa\_v1}), we apply the GaLore optimization only to the attention parameters of the image encoder for fine-tuning and the prompt encoder and mask decoder are fine-tuned without using GaLore. Even though GaLore is only applicable to the immediate attention layer succeeding MLP layer of the transformer neural network, we apply it on all the attention parameters in ViT of SAM to justify the same. In the second variation (referred to as \textit{MedSAGa\_v2}), we apply GaLore to all the parameters of the image encoder and fine-tune it and do not perform any fine-tuning on the prompt encoder and mask decoder. Both these variations were tested on the Chest X-ray dataset and the results of MedSAGa\_v1, MedSAGa\_v2 along with the best-performing model, MedSAGa are presented in Table~\ref{tab:variation-table}. Our second category of ablation studies is based on experimenting with the effects of applying warmup while fine-tuning MedSAGa. Fig.~\ref{fig:my-figure} shows the results of segmentation performance on all the datasets when fine-tuned with and without warmup. The results of the same study on other datasets is mentioned in the supplementary material. 

\section{Limitations and Future Scope}
The MedSAGa methodology leverages a large-scale model akin to SAM for few-shot image segmentation within resource-constrained environments, exhibiting a marked reduction in memory utilization compared to other SOTA models, while still yielding comparable segmentation performance. However, attaining optimal segmentation performance with MedSAGa may not always be feasible, as varying architectural designs may excel in capturing the distinctive features of diverse datasets. Additionally, as highlighted in the GaLore literature, further improvements in memory overhead can be achieved by reducing the projection layer dimensions via techniques such as quantization and streamlined parameterization, which can be done as part of the future work. Since this work presents SAM in culmination with GaLore (a gradient low-rank optimization technique) it can be applied to various end-to-end training techniques of models involving ViTs. Not just image segmentation, the use of GaLore alongside ViT can be interestingly used for a wide-variety of downstream tasks in a resource-constraint environments.

\section{Conclusion}
In our work, we demonstrate the use of large-scale models like SAM for few-shot medical image segmentation in resource-constrained settings like healthcare by using Gradient low-rank projection (GaLore) for fine-tuning the image encoder. This allows us to achieve significant memory efficiency while still utilizing full parameter training. Our rigorous experiments on diverse medical image segmentation datasets showcase our approach's effectiveness in resource-constrained environments. Easy integration with SAM architecture proves the practicality and efficiency of MedSAGa for healthcare implementation.

{
    \small
    \bibliographystyle{ieeenat_fullname}
    \bibliography{main}
}


\end{document}